\documentclass{article}

\usepackage{arxiv}

\usepackage[utf8]{inputenc} 
\usepackage[T1]{fontenc}    
\usepackage{hyperref}       
\usepackage{url}            
\usepackage{booktabs}       
\usepackage{amsfonts}       
\usepackage{nicefrac}       
\usepackage{microtype}      
\usepackage{lipsum}
\usepackage{graphicx}
\usepackage{natbib}
\usepackage{amsmath}
\usepackage{multirow}
\usepackage{float}

\title{Typoglycemia under the Hood: \\  Investigating Language Models’ Understanding of Scrambled Words}

\author{
 Gianluca Sperduti \\
 Institute of Information Science and Technologies\\
 National Research Council\\
 University of Pisa\\
 56124 Pisa, Italy \\
 \texttt{gianluca.sperduti@isti.cnr.it} \\
 \And
 Alejandro Moreo \\
 Institute of Information Science and Technologies\\
 National Research Council\\
 56124 Pisa, Italy \\
 \texttt{alejandro.moreo@isti.cnr.it} \\
}

\begin{document}
\maketitle
\begin{abstract}
Research in linguistics has shown that humans can read words with internally scrambled letters, a phenomenon recently dubbed ``typoglycemia''. 
Some specific NLP models have recently been proposed that somehow similarly demonstrate robustness to such distortions by ignoring the internal order of characters by design. 
This raises a fundamental question: how can models perform well when many distinct words (e.g., ``form'' and ``from'') collapse into identical representations under typoglycemia? Our work, focusing exclusively on the English language, seeks to shed light on the
underlying aspects responsible for this robustness.
We hypothesize that the main reasons have to do with the fact that (i) relatively few English words collapse under typoglycemia, and that (ii) collapsed words tend to occur in contexts so distinct that disambiguation becomes trivial. In our analysis, we (i) analyze the British National Corpus to quantify word collapse and ambiguity under  typoglycemia, (ii) evaluate BERT’s ability to disambiguate collapsing forms, and (iii) conduct a probing experiment by comparing variants of BERT trained from scratch on clean versus typoglycemic Wikipedia text;
our results reveal that the performance degradation caused by scrambling is smaller than expected.
\end{abstract}

\section{Introduction}
\noindent
Humans’ ability to comprehend words with internally jumbled letters—known as the \emph{typoglycemia} phenomenon—has long attracted attention from linguists, psychologists, and the general public \cite{grainger2004does,jumbledwords}. A widely circulated (and misattributed) Internet meme, purportedly from a Cambridge University study,\footnote{The meme states: \emph{``Aoccdrnig to a rscheearch at Cmabrigde Uinervtisy, it deosn't mttaer in waht oredr the ltteers in a wrod are, the olny iprmoetnt tihng is taht the frist and lsat ltteer be at the rghit pclae. The rset can be a toatl mses and you can sitll raed it wouthit porbelm. Tihs is bcuseae the huamn mnid deos not raed ervey lteter by istlef, but the wrod as a wlohe.''}} claims that as long as the first and last letters of a word remain in place, the order of the internal letters has little effect on reading comprehension. Although anecdotal, this aligns to some extent with psycholinguistic evidence indicating that readers can tolerate internal character rearrangements, though not without some cost \cite{healy1976detection,mccusker1981word,andrews1996lexical,marian2012clearpond}.

Interestingly, this ability is not exclusive to \emph{humans}. Prior work has shown that typoglycemia can be transferred to NLP models through representations based on unordered bag-of-characters  \cite{SakaguchiDPD17}, internal-character averaging \cite{belinkovB18}, or training with reordered text \cite{sperduti2021garbled}. More recently, \citet{wang2025} demonstrated that large language models (LLMs) also display some degree of resilience to such distortions.
This resilience is surprising from an explainability perspective, as ignoring internal character order can cause semantically distinct English words—such as \emph{form} and \emph{from}—to collapse into the same representation, potentially leading to ambiguity and substantial information loss. Yet empirical findings indicate only a modest performance drop. 

While previous work has primarily focused on developing computational mechanisms to enhance resilience to order-based misspellings, we are not aware of dedicated studies that aim to shed light on the distributional characteristics underlying this phenomenon.
In this work, we therefore seek to advance the understanding of typoglycemia through the following analyses:

\begin{description}
    \item [Quantification and characterization of the collapsing word problem:] using the British National Corpus \cite{bnc2007}, we measure the frequency and distribution of collapsing words—distinct words that become indistinguishable after internal characters are reordered—under two scenarios of typoglycemia: \emph{classic typoglycemia} (in which only internal word letters are sorted) and \emph{extreme typoglycemia} (in which the full set of word letters are alphabetized) (\S\ref{sec:collapsing_quant}).
    
    \item[Probing contextual disambiguation:] we use BERT \cite{bert} as a proxy in a word-disambiguation experiment to test whether the correct collapsing word can be recovered from masked-sentence contexts (for instance, whether BERT is able to identify the word \emph{from} as more likely than the collapsing alternative \emph{form} in the sentence \emph{I am originally [MASK] New York}). We test this for both typoglycemia settings, showing that BERT almost always selects the correct word based on contextual cues (\S\ref{sec:disambiguation}).

    \item[Probing the feasibility of typoglycemia-based pretraining for language models:] we train three mini BERT models from scratch on clean, classic-typoglycemic, and extreme-typoglycemic English Wikipedia text, and evaluate them on GLUE \cite{glue}, with the aim of quantifying the drop in performance that results from  typoglycemic inputs (\S\ref{sec:typoglycemic_modeling}).

\end{description}

Our analysis suggests that the main reasons why typoglycemic text can still be successfully processed may be related to the fact that (i) the prevalence of collapsing words is relatively low in English, and (ii) collapsing words generally convey distinct meanings, making disambiguation by context relatively straightforward.

The rest of this paper is structured as follows. In Section~\ref{sec:related} we discuss related work. In Section~\ref{sec:typoglycemic} we analyze the collapsing phenomenon and the BERT's ability to disambiguate between collapsing words. In Section~\ref{sec:typoglycemic_modeling}, we report on experiments where we train BERT on differently scrambled versions of Wikipedia text. 
Section~\ref{sec:conclusions} offers some concluding remarks while Section~\ref{sec:limitations} discusses some limitations of our work.

\section{Related work}
\label{sec:related}

\subsection{NLP models' robustness to misspellings}

Since typoglycemic models are often developed to gain greater resistance to misspellings,
it is relevant to include some discussion of NLP robustness to misspellings. 
In NLP, model robustness to corrupted text has been evaluated extensively. \citet{heigold-etal-2018-robust} find that common neural models degrade significantly under text noise, though training on noisy data can mitigate this. Even transformers-based contextual models struggle with corrupted input \cite{yang2019can,kumarMG20,moradi,RavichanderDRMH21,satheesh}. \citet{nguyen2020word} show that SGNS outperforms \texttt{fastText} in most intrinsic tasks, though \texttt{fastText} fares better with deleted characters. Both models perform well on emotion-rich, intentionally misspelled words (e.g., the elongated form ``sooorry'').
More recently, many tests on misspellings and noise have also been carried out with Large Language Models (LLMs). In particular, the study by \citet{ebench} tested models such as LLaMA, Vicuna, GPT-3.5, and GPT-4 against various types of noise. \citet{panLX24}, instead, evaluated the robustness of large language models (LLMs) to misspellings for the machine translation task. Among all models, GPT-4 appears to be the most robust and resilient, approaching human abilities in resilience to misspellings.

\subsection{Typoglycemia and NLP} 

Several studies have investigated how the internal arrangement of characters within words impacts NLP. 
One of the earliest contributions in this area is by \citet{SakaguchiDPD17}, who introduced the \emph{Semi-Character Recurrent Neural Network} (ScRNN). This model treats the initial and final characters of a word as distinct one-hot vectors, while encoding the internal characters as an unordered collection—a bag-of-characters—effectively ignoring their sequence. ScRNN was designed for spelling correction tasks rather than broader NLP applications. \citet{PruthiDL19} later incorporated ScRNN as the first component in a two-stage framework, demonstrating that ScRNN can improve resistance to different types of misspellings for word error correction and sentiment classification.
Subsequently, \citet{belinkovB18} presented a different approach called \emph{meanChar} for machine translation. This method computes a word representation by averaging its character embeddings and processes the result using a word-level encoder, similar in spirit to the CharCNN model by \citet{Kim14charcnn}.

Another noteworthy method is \emph{Robust Word Vectors} (RoVe), introduced by \citet{MalykhLK18}. RoVe creates three distinct vectors for each word: \emph{Begin} (B), \emph{Middle} (M), and \emph{End} (E). These vectors are derived from the sum of one-hot encodings of specific character segments. For example, in the word \emph{previous}, B corresponds to the first three characters (\emph{pre}-), E to the final three (-\emph{ous}), and M to all the other characters in the word (-\emph{vi}-). RoVe demonstrated strong performance across multiple languages, including English, Russian, and Turkish, and was evaluated on tasks such as paraphrase detection, sentiment analysis, and textual entailment recognition. RoVe shows a good level of resilience to different types of misspellings (character insertion, deletion, etc.).
\citet{sperduti2021garbled} showed that static embedding models such as Word2Vec can retain strong performance when trained on BE-sorted data—where internal letters are alphabetically ordered while preserving the first and last characters in place— and on FULL-sorted data—where all letters are alphabetically sorted. In our study, we apply the same sorting procedures to all words in the British National Corpus to investigate the impact and implications of word collapse when character order is disregarded.

Recent studies have examined the robustness of large language models (LLMs) to scrambled or noisy input. \citet{CaoKMI23} demonstrated that models such as GPT-4 \cite{gpt4} exhibit substantial resilience to garbled text, even when character sequences are heavily disrupted. Similarly, \citet{wang2025} introduced a new evaluation metric, \emph{SemRecScore}, and showed that word shape plays a more critical role than context in determining LLMs' robustness to scrambled input.
Trained on large datasets —likely including some natural misspellings— these models seem to draw on patterns in text rather than on explicitly designed cognitive processes.

In contrast, our work addresses a complementary question: can typoglycemia be incorporated into transformer-based contextual models during training itself, if they are trained from scratch on typoglycemic data? How is this possible, given the challenge of collapsing words? This distinction is critical, as inference-time robustness does not necessarily imply that models can acquire the human-like ability to read scrambled text by design.
Our approach emphasizes a lightweight, controlled setup, grounded in cognitive and theoretical motivations rather than scale.  

\section{Typoglycemia in NLP: Why does it work?}
\label{sec:typoglycemic}

We start by analyzing the possible factors that make the problem tractable under typoglycemic conditions.
In Section~\ref{sec:method_b2v}, we describe different techniques for altering word order.
Section~\ref{sec:hypotheses} introduces our working hypotheses and outlines how we validate them.
In Section~\ref{sec:collapsing_quant}, we present the approach used to quantify the extent to which our hypotheses hold, and in Section~\ref{sec:disambiguation} we propose and carry out a disambiguation experiment.

\subsection{Sorting, \emph{sinortg}, \emph{ginorst}: A simple NLP method to explore typoglycemia}
\label{sec:method_b2v}

To simulate typoglycemia, we adopt a lightweight pre-processing method which corresponds to the approach of our previous work \cite{sperduti2021garbled}.  
Given a word
\[
w = [c_1, c_2, \ldots, c_n],
\]
where \( c_i \) denotes the character at position \( i \), we apply a transformation called \textbf{BE-sorting}, which corresponds to \textbf{classic typoglycemia} in our experiments.
The transformation produces an artificial token defined as:
\[
R_{\text{BE}}(w) = [c_1, \mathrm{sort}([c_2, \ldots, c_{n-1}]), c_n]
\]
In other words, the first (\textbf{B}eginning) and last (\textbf{E}nd) characters of each word are preserved, while the middle characters (all except the first and last) are rearranged in alphabetical order. 
For example, for the word \texttt{embedding}  
(\([e, m, b, e, d, d, i, n, g]\)),  
the BE-sorted version is \texttt{ebddeimng}.  
Any scrambled form of \texttt{embedding} that keeps the same first and last letters ---such as \texttt{ebdimnedg} or \texttt{edmnbdieg}--- will map to the same BE-sorted token \texttt{ebddeimng}.
Note that BE-sorting only has an effect on words with at least 4 characters.

Conversely, \textbf{Full-sorting} alphabetically reorders all characters, including the first and last:
\[
R_{\text{Full}}(w) = \mathrm{sort}([c_1, c_2, \ldots, c_n])
\]
For the word \texttt{embedding}, this yields \texttt{bddeeimmng}, which discards positional information entirely. This approach corresponds to \textbf{extreme typoglycemia} in our experiments. Note that FULL-sorting affects words with at least 2 characters.

\subsection{This works, but how? The problem of the collapsing effect}
\label{sec:hypotheses}

As stated in Section~\ref{sec:related}, several methods use the cognitive human concept of typoglycemia and develop it into NLP models. 
However, theoretically, these models should lose a lot of information due to a collapse of words that occurs when the order of the characters in the words is ignored: \textbf{the collapsing effect}.
The collapsing effect occurs:

\begin{itemize}

\item With \textbf{classic typoglycemia} when a word has the beginning and ending letters identical to another word and also shares the same internal characters, but in a different order. An example is the pair \textit{salt} and \textit{slat}.

\item With \textbf{extreme typoglycemia} when a word has the same characters as another word, e.g., the words \textit{listen} and \textit{silent} share exactly the same characters but in a different order, making them indistinguishable under extreme typoglycemia.

\end{itemize}

The collapsing effect should lead to a significant loss of semantics, as two distinct words would collapse into a single representation (some examples are provided in Table \ref{tab:collapsed_examples}).
This raises the question of how some NLP models are nevertheless able to effectively learn from typoglycemic text.
We start our exploration following two working hypotheses:

\begin{enumerate}

\item Words that collapse are relatively rare in English.

\item Words that collapse  tend to appear in very different contexts, so distributional semantic-based models might ultimately be able to distinguish the intended term.

\end{enumerate}

To examine whether our hypotheses hold, we analyze how many words collapse in the British National Corpus and their average frequency. To this end, we uses BERT as a proxy to assess how easily a collapsed word can be recognized from its context.

\subsection{Quantifying Collapsing Effects in Typoglycemia}
\label{sec:collapsing_quant}

\noindent We used the British National Corpus (BNC) as our reference dataset, given its extensive use in prior linguistic studies for statistical analysis of English \cite{adjective_english,significant_or_random}.
The analysis proceeds as follows: (i) we tokenize by splitting on whitespaces the BNC, resulting in a vocabulary of 1,094,782 unique words, and (ii) we then apply the sorting procedures described in Section~\ref{sec:method_b2v} to each word in the vocabulary.
We distinguish between two related notions:
\begin{itemize}

    \item (Pre-)collaps\textbf{ing} words (\emph{\#ing-ws}) are the \textbf{original words} from the vocabulary that map onto a given collapsed form.
    
    \item (Post-)collaps\textbf{ed} words (\emph{\#ed-ws}) are the \textbf{resulting forms} obtained after applying a typoglycemic transformation to two or more original collapsing words. 
    They act as \emph{``umbrella forms''} potentially representing several distinct words in their scrambled state.

\end{itemize}

\noindent In other words, several distinct original words (collapsing words) can merge into the same scrambled form (collapsed word).

\smallskip
\noindent\textbf{Example.}  
Consider the words \textit{form} and \textit{from} under \emph{classic typoglycemia} (scrambling internal letters only);  
in this case:

\begin{itemize}
    \item \textit{form} and \textit{from} (original) are the \textbf{collapsing words}.
    
    \item \textit{form} (scrambled) is, at the same time, the \textbf{collapsed word}.    
\end{itemize}

\noindent Table~\ref{tab:collapsed_examples} provides additional examples of this mapping.

\begin{table}[tb]
\centering
\caption{Examples of collapsed (\emph{ed-ws}) and collapsing (\emph{ing-ws}) words.}
\begin{tabular}{l|l|l}
\hline
\textbf{Type of typoglycemia} & \textbf{\#ed-ws (umbrella)} & \textbf{\#ing-ws}                \\ \hline
\multirow{5}{*}{Classic (BE-sorting)}        & tehre                               & there, three      \\ 
~ & radeinrwg & rewarding, redrawing \\
~ & part & prat, part \\
~ & ecepxt & except, expect \\
~  & corsue & Crusoe, course \\
\hline
\multirow{5}{*}{Extreme (Full-sorting)}      & ceeilns                            & silence, license \\ 
~      & acesu                            & sauce, cause \\ 
~      & how                            & who, how, woh \\ 
~     & hitw                            & with, whit \\ 
~     & almnor                            & Marlon, normal \\\hline 
\end{tabular}
\label{tab:collapsed_examples}
\end{table}

\begin{figure}[tb]
    \centering
    \includegraphics[width=0.65\textwidth]{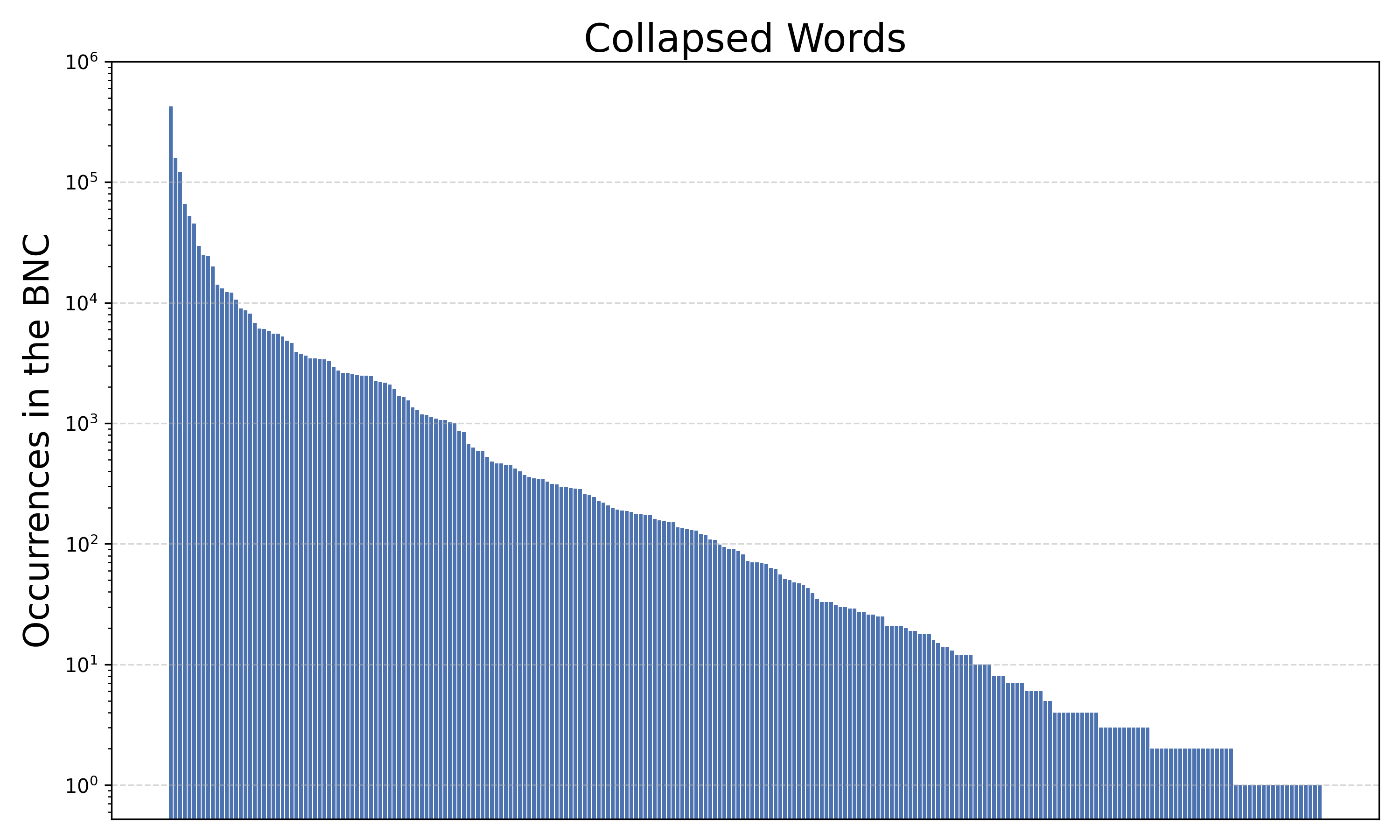}
    \caption{Frequency distribution of collapsing words in the classic typoglycemia setting (log scale).}
    \label{fig:bar_of_frequency_BE}
\end{figure}

\begin{figure}[tb]
    \centering
    \includegraphics[width=0.65\textwidth]{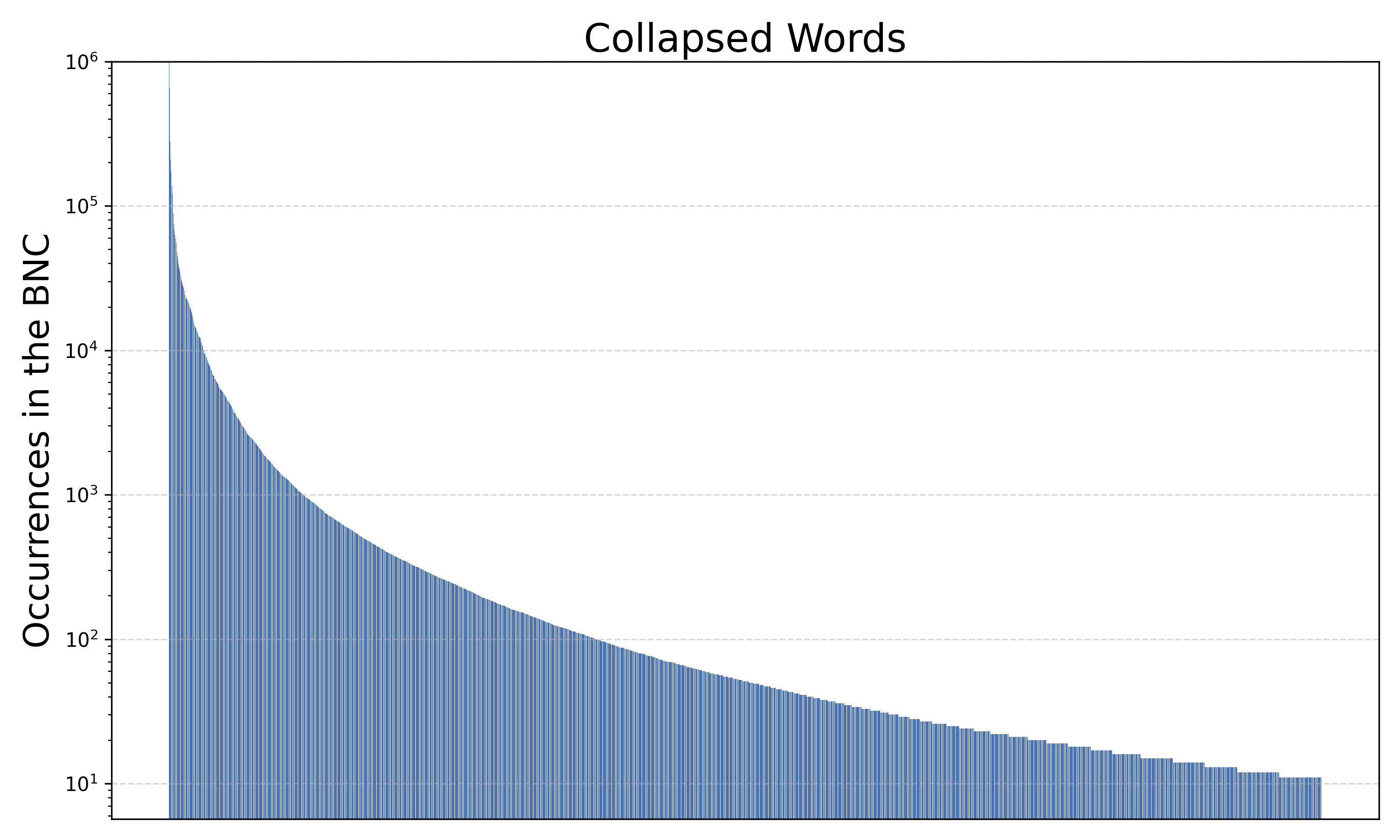}
    \caption{Frequency distribution of collapsing words in the extreme typoglycemia setting (log scale).}
    \label{fig:bar_of_frequency_full}
\end{figure}

Processing the BNC vocabulary yielded the statistics reported in Table~\ref{tab:typoglycemia} for classic typoglycemia and extreme typoglycemia, while Figures~\ref{fig:bar_of_frequency_BE} and~\ref{fig:bar_of_frequency_full} show the frequency distributions (in log scale) for the collapsing words in both settings, respectively.

\begin{table}[tb]
    \centering
    \caption{Collapsed and collapsing word statistics for the BNC under classic and extreme typoglycemia. Specifically, we report the number of unique collapsed words (\#ed-ws) and collapsing words (\#ing-ws), the average frequency of collapsing words (Avg.), and the standard deviation of the frequency of unique collapsing words (Std.).}
    \begin{tabular}{l|c|c|c|c}
        \hline
        \textbf{Typog. style} & \textbf{\#ed-ws} & \textbf{\#ing-ws} & \textbf{Avg.} & \textbf{Std.} \\
        \hline
        Classic  & 681 & 1,372   & 1,828 & 16,023 \\
        Extreme  & 5,703 & 13,904 & 2,567  & 59,326 \\
        \hline
    \end{tabular}
    \label{tab:typoglycemia}
\end{table}

The portion of the vocabulary corresponding to collapsing words is relatively small compared to the total BNC vocabulary (approximately 0.12\% for the classic setting; approximately 1.27\% for the extreme setting), for more detailed on the number of unique words identified numbers see Table \ref{tab:typoglycemia}.
Average frequencies are higher for the extreme variant, but the large standard deviation (Table~\ref{tab:typoglycemia}) indicates substantial skew, also visible in Figure~\ref{fig:bar_of_frequency_full}. Standard deviation refers to the dispersion of word frequencies in the BNC for the identified unique collapsing words from the vocabulary, indicating a highly skewed distribution with a few high-frequency outliers.  
One possible contributing factor is that extreme typoglycemia also produces collapsed forms for words with fewer than four characters---which correspond to approximately 44\% of all words in the corpus---, whereas in the classic setting such words remain distinct because at least two internal characters must remain fixed.
While this analysis is purely corpus-based, the relatively small proportion of collapsing words and the observed frequency distributions provide initial insight into \textit{why} the collapsing effect may not pose a significant problem in practice.  
To further validate this intuition, we conduct a disambiguation experiment in the next section, aiming to confirm that these collapsed words tend to appear in distinct contextual environments.

\subsection{The proof is in the pudding: collapsing words are easy to tell apart using context}
\label{sec:disambiguation}

\noindent Corpus-based findings show that collapsing words are rare.
Our second goal is therefore to analyze the contexts in which these words appear.
We hypothesize that NLP models can read scrambled text because possible collapsing words are easy to distinguish through context.

Let us return to our example of \textit{form} and \textit{from} again. Although the collapsed word coincide, the collapsing words carry very very distinct meanings, making it easy to disambiguate the correct one in context; e.g., in the sentence ``I have just signed the (\emph{form?/from?})''. If our conjecture is correct, we would therefore expect a contextualized language model to be able to  chose the right choice more often than not.

To test this hypothesis, we use BERT in a word disambiguation experiment. Unlike standard disambiguation tasks that aim to predict the most likely meaning given a word in context, our task is to identify the preferred collapsing word given a context including a collapsed word.

We chose base BERT (uncased) as a reference model due to its widespread use and given that we know that the pre-trained model available has not been trained on the BNC, thus ensuring our analysis is not contaminated. We deliberately avoided fine-tuning to evaluate BERT's out-of-the-box contextual understanding, using the original masked language modeling objective.

In particular, we created the disambiguation task as follows: We first extracted all BNC sentences containing collapsing words for both the \emph{classic} and \emph{extreme} typoglycemia variants. For each sentence, we replaced the collapsed word, i.e. the umbrella term, with a single \texttt{[MASK]} token to create a prediction instance. We then evaluate the extent to which BERT assigns a higher probability to the correct collapsing word (as appearing in the original sentence) than to the other(s) collapsing words. In other words, given a set of real lexical candidates that would collapse to the same scrambled form, we measure how often BERT can correctly recover the intended word from context. 
Note that the experiment is meant to validate our hypothesis that collapsing words tend to occur in fairly dissimilar contexts, and is not meant to evaluate the ability of BERT to work under typoglycemic conditions; indeed, note that the sentence serving as context is presented in its clean form (i.e., without scrambling).
A schematic example of this disambiguation setup is shown in Table~\ref{tab:disamb_example}.

\begin{table}[tb]
\caption{Example from the disambiguation experiment for classic typoglycemia. Note that some sentences in the BNC are ungrammatical or dialectal.}
\centering
\resizebox{\columnwidth}{!}{
\begin{tabular}{l|l}
\hline

\textbf{Original sentence}   & This chestnut was second to Royal \textbf{Cedar} at Newbury last month.  \\\hline

\textbf{Collapsed word}        & \emph{acder} \\\hline

\textbf{Collapsing words} (choices)    & \emph{cedar}, \emph{cader} \\\hline

\textbf{Masked sentence}     & This chestnut was second to Royal \textbf{[MASK]} at Newbury last month. \\\hline

\textbf{Correct collapsing word}        & [MASK]=\emph{cedar} \\\hline

\end{tabular}}
\label{tab:disamb_example}
\end{table}

For each masked sentence:
\begin{itemize}
    \item We obtain the model's output logits corresponding to the \texttt{[MASK]} position for all tokens in the vocabulary. 
    \item For each candidate (collapsing) word, we extract the logits of all its subword tokens according to BERT's WordPiece tokenizer. If a word consisted of multiple tokens (e.g., ``playing'' → ``play'', ``\#\#ing''), we averaged their logits to obtain a single scalar score representing the model’s confidence in that word.
    \item We then compare these scores across all candidates and select the word with the highest average logit as BERT's predicted choice for that context.
    
\end{itemize}

\noindent The results are shown in  Table~\ref{tab:disamb_experiment}. This table reveals that BERT almost always preferred the correct collapsing word, both in  the \emph{classic} and \emph{extreme} typoglycemia setups, suggesting that local context alone is often sufficient to disambiguate collapsing words.

We also checked whether there was a correlation between word frequency and disambiguation accuracy.
The analysis shows a very weak positive Pearson correlation between word frequency and BERT accuracy in both classic (0.099) and extreme (0.044) typoglycemia, thus suggesting performance is largely independent of word frequency.

\begin{table}[tb]
\caption{Accuracy in the disambiguation experiment for BE-sorted and fully sorted variants. \textbf{\#sentences} represent the number of sentences that included collapsing words for each of the variants, which is the dimension of the dataset. More details on the disambiguation experiment can be found in Appendix \ref{sec:disamb_examples}. } 
\centering
\begin{tabular}{ccc}
\hline
\textbf{Approach} & \textbf{\#sentences} & \textbf{Accuracy} \\
\hline
Classic typoglycemia            & \phantom{0}2,551,690         & 96.58\% \\
Extreme typoglycemia          & 36,231,306        & 97.01\% \\
\hline
\end{tabular}
\label{tab:disamb_experiment}
\end{table}

Given these findings that help explain why typoglycemic models work in the NLP world, we turned to a new question: Can a language model be trained from scratch on typoglycemic text, and if so, how well do those abilities transfer? This is the topic of the following section.

\section{Typoglycemic modeling: contextual model can learn from typoglycemic text}
\label{sec:typoglycemic_modeling}

Through our analysis of the collapsing problem, we observed that the potential information loss from typoglycemic text is likely limited. This suggests that, in principle, contextual models trained on such text could still achieve competitive performance. Training on typoglycemic text would produce models that, by design, process words without relying on internal character order.
To test this, we conducted a feasibility study by training three miniature masked language models based on the BERT architecture: one on the original (unsorted) English Wikipedia, one on a version processed with the BE-sorted technique (classic typoglycemia), and one on a fully sorted version (extreme typoglycemia). Our aim is not to produce a novel BERT variant, but rather to probe whether typoglycemia can be integrated into contextual models by design.
We evaluated all models on the GLUE benchmark to assess whether typoglycemia-trained BERT variants acquire transferable linguistic competence.

\subsection{Models and Training Details}
\label{sec:models_details}

Each model was trained for 500{,}000 steps (approximately 20 hours) with a batch size of 16 and evaluation every 2{,}000 steps (other details can be found in Appendix \ref{sec:model_hyp}). Models are trained from scratch, and the training corpus consisted solely of the March 2022 English Wikipedia snapshot (\texttt{20220301.en}).
The models include:

\begin{description}

    \item[Clean BERT (C-BERT):] Trained on the unmodified Wikipedia corpus. Serves as a baseline using the same data, hyperparameters, and training time as the typoglycemia variants.

    \item[Classic Typoglycemic BERT (T-BERT):] Trained on Wikipedia preprocessed with the BE-sorted technique. 
    
    \item[Extreme Typoglycemic BERT (X-BERT):] Trained on Wikipedia preprocessed with extreme typoglycemia.
\end{description}

\subsection{Tokenizer Construction and Analysis}
\label{sec:tokenizer}
\noindent

To train the three models, we built tokenizers from scratch using the Wikipedia corpus, following the WordPiece approach employed in the original BERT and maintaining the standard vocabulary size of 30{,}522 tokens. The first tokenizer corresponds to the standard (classic) variant, while the other two were derived from the classic typoglycemia and extreme typoglycemia preprocessing strategies, respectively. For model training, we apply word corruption only to words of length  equal or greater than four in both cases, in order to facilitate a direct comparison between the models.

\begin{figure}
    \centering
    \caption{Token frequency distributions (log scale) for the tokenizers in the classic and typoglycemic variants.}
    \includegraphics[width=0.75\textwidth]{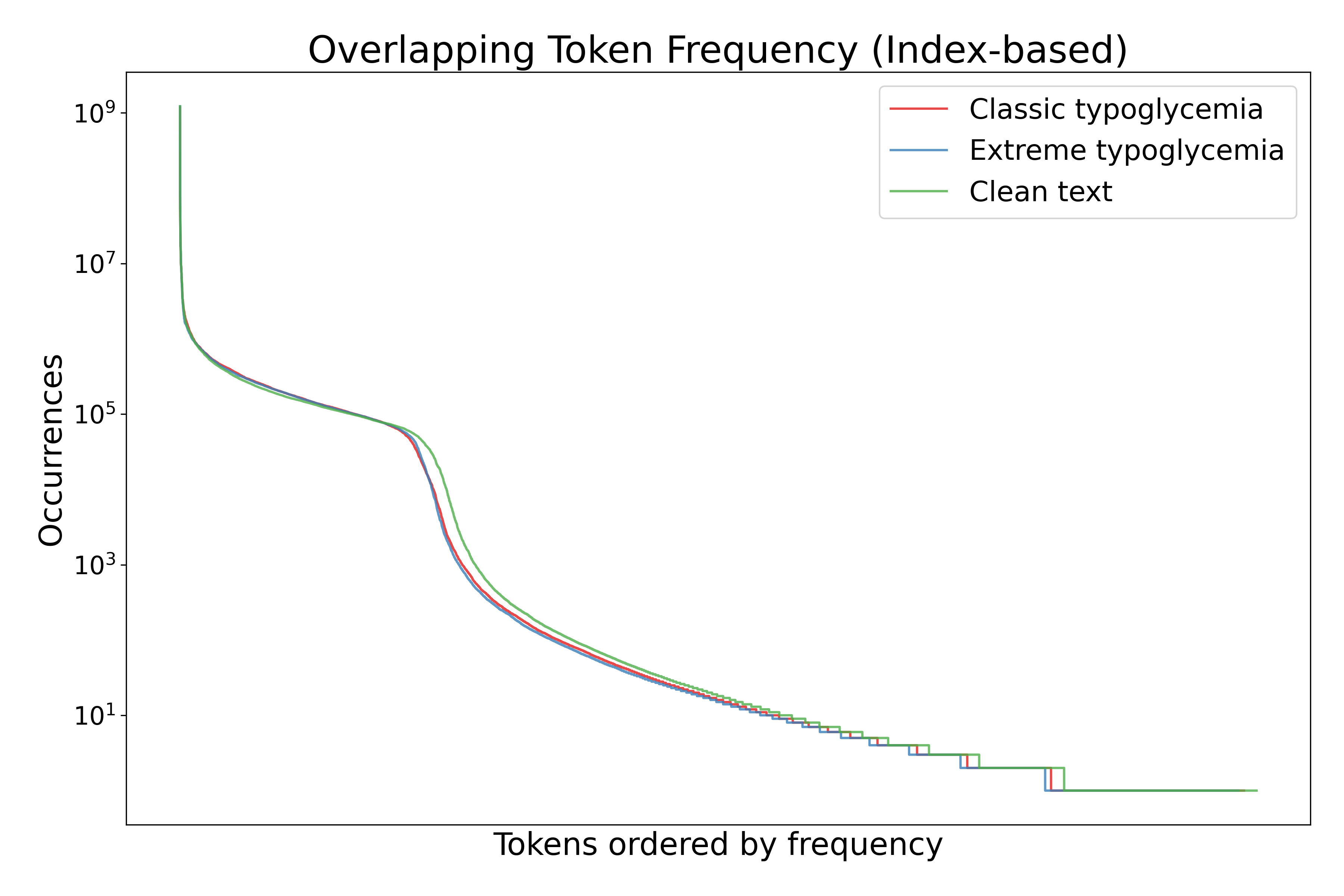}
    \label{fig:garbled_tokenizers.png}
\end{figure}

Token frequency distributions for the three tokenizers are presented in Figure~\ref{fig:garbled_tokenizers.png} (log scale). All three models exhibit the characteristic long-tailed distribution, in which a small subset of tokens occurs very frequently while the vast majority are relatively rare. Relative to the classic tokenizer, both typoglycemic variants display a shift in the distribution, indicating a higher concentration of high-frequency tokens and a slightly reduced proportion of rare tokens. The effect is slightly more pronounced for the extreme typoglycemic tokenizer, which shows a steeper skew, though this may partly reflect the influence of preprocessing rather than a true reduction in vocabulary diversity. Further details about the tokenizers can be consulted in Appendix \ref{sec:more_on_tokenizer}. 

Importantly, it should be noted that, due to tokenization, BERT (as well as other subword-based models) do not have direct access to the internal structure of individual words. For instance, if the word \emph{person} is tokenized as a single unit, its scrambled variant \emph{peorsn} is treated as a completely distinct token, and the model cannot leverage letter-level relationships as humans do when reading typoglycemic text. This implies that, for the model, typoglycemia may effectively correspond to presenting an arbitrary new token rather than a corrupted version of a familiar word. Consequently, the impact of typoglycemia in this setting primarily reflects the model's ability to 
capture underlying distributional patterns of meaning, even though some of the contextual words may have collapsed,
and thus has nothing to do with
a human-like cognitive resilience to internal letter reordering.
Despite this limitation, we consider it an open and interesting question whether such preprocessing might allow for some empirical transfer of this human-like robustness. To see a more direct example of our tokenization, see Table \ref{tab:tokenized_sentence_examples} in Appendix \ref{sec:more_on_tokenizer}. 

\subsection{GLUE Benchmark Evaluation}
\label{sec:glue_results}

We evaluated the three models described in Section~\ref{sec:models_details} on the GLUE benchmark suite. 
The three models (C-BERT, T-BERT, X-BERT) were fine-tuned on each GLUE task without additional hyperparameter tuning (see Appendix \ref{sec:more_on_tokenizer}for further details). 
The results we have obtained are reported in Table~\ref{tab:bert_glue_scores}.

\begin{table}[H]
\centering
\caption{GLUE benchmark scores for our three models. Bold values indicate the best score per task, excluding the benchmark model.}
\resizebox{0.5\columnwidth}{!}{
\begin{tabular}{lccc}
\hline
\textbf{Task} & \textbf{C-BERT} & \textbf{T-BERT} & \textbf{X-BERT} \\
\hline
QQP    & \textbf{0.852} & 0.824 & 0.844 \\
MRPC   & 0.831 & 0.839 & \textbf{0.840} \\
CoLA   & \textbf{0.195} & 0.106 & 0.072 \\
QNLI   & \textbf{0.855} & 0.817 & 0.835 \\
STS-B  & \textbf{0.851} & 0.827 & 0.827 \\
SST-2  & 0.847 & 0.785 & \textbf{0.858} \\
MNLI   & \textbf{0.771} & 0.707 & 0.754 \\
RTE    & 0.498 & 0.494 & \textbf{0.519} \\
\hline
Avg    & \textbf{0.712} & 0.675 & 0.694 \\
\hline
\end{tabular}}
\label{tab:bert_glue_scores}
\end{table}

Under identical training conditions, C-BERT achieved the highest average score among our models (0.712), surpassing T-BERT by 0.037 and X-BERT by 0.018. Both typoglycemic variants still demonstrated relatively high performance: T-BERT obtained an averaged score of 0.675, while X-BERT slightly outperformed it with an averaged score of 0.694, despite employing more aggressive character sorting. Overall, the findings indicate that typoglycemic preprocessing—even in its extreme form— does not prevent large contextual models from learning effective representations. However, performance on CoLA (The Corpus of Linguistic Acceptability---a task evaluating the grammatical acceptability of English sentences) \cite{cola}, was notably low across all models, likely reflecting the reduced training budget and absence of hyperparameter tuning, both of which may have constrained performance on syntax-sensitive tasks. In particular, performance on CoLA was disastrous for the typoglycemic models, indicating a possible detrimental impact of typoglycemia modeling on syntactic and grammatical learning.

\subsection{Does Extended Training Help? Evaluating typoglycemic BERTs with Additional Steps}
\noindent

We investigated whether increasing the number of training steps could further improve the performance of the typoglycemic models. While it is generally expected that longer training leads to better language model performance, it is not obvious that this holds for models trained on modified text, such as sorted (typoglycemic) inputs. Instead, it could well be the case that the limitations we have witnessed rather stem from the fact that some information is irretrievably lost during the scrambling process. This hypothesis stems from the fact that sorting leads to a form of obfuscation: if the information is still there, additional training epochs might suffice for the model to develop comparable representational capacity.

\begin{figure}
    \centering
    \includegraphics[width=0.5\linewidth]{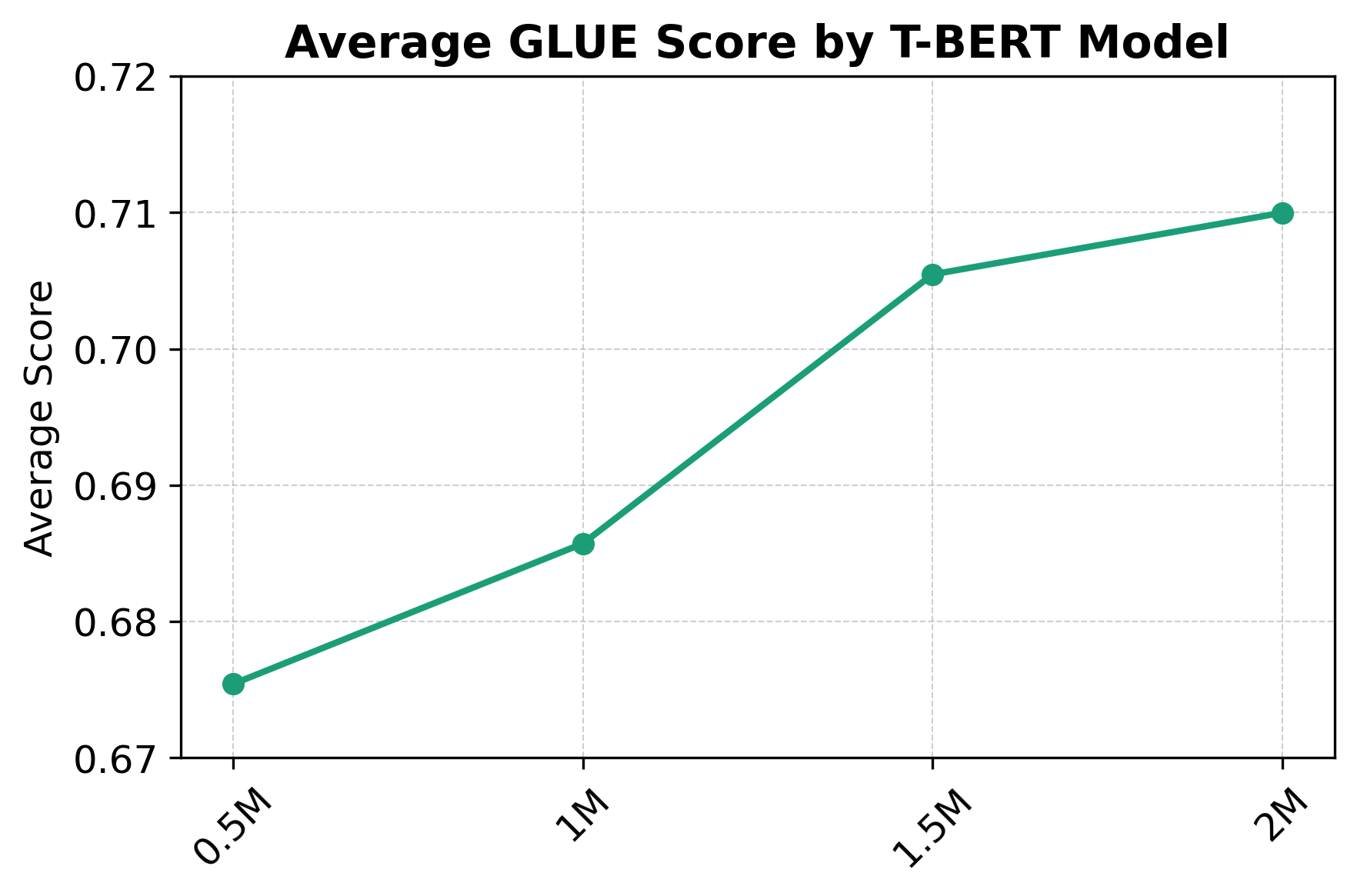}
    \caption{Growth of the average GLUE score of T-BERT variants as training steps increase.}
    \label{fig:TBERT_grown}
\end{figure}

As shown in Table~\ref{tab:typbert_adding_steps_transposed} and Figure~\ref{fig:TBERT_grown}, extending the training phase significantly improved T-BERT's performance. At 2 million steps, the model's average GLUE score nearly matches that of C-BERT (reported in Section~\ref{sec:glue_results}), confirming our hypothesis. 

\begin{table}[tb]
\centering
\caption{GLUE scores for T-BERT with increasing training steps. Bold indicates the best result overall, while the best results obtained by T-BERT are underlined.}
\begin{tabular}{lccccc}
\hline
Model & C-BERT & T-BERT & & & \\
\hline
\textbf{Task} & \textbf{0.5M} & \textbf{0.5M} & \textbf{1M} & \textbf{1.5M} & \textbf{2M} \\
\hline
\textbf{QQP}   & 0.852 & 0.824 & 0.834 & 0.852 & \underline{\textbf{0.855}} \\
\textbf{MRPC}  & 0.831 & 0.839 & 0.813 & 0.818 & \underline{\textbf{0.847}} \\
\textbf{CoLA}  & \textbf{0.195} & 0.106 & 0.141 & \underline{0.165} & 0.138 \\
\textbf{QNLI}  & \textbf{0.855} & 0.817 & 0.829 & 0.842 & \underline{0.848} \\
\textbf{STS-B} & \textbf{0.851} & 0.827 & 0.826 & 0.835 & \underline{0.838} \\
\textbf{SST-2} & \textbf{0.847} & 0.785 & 0.814 & \underline{0.846} & 0.842 \\
\textbf{MNLI}  & \textbf{0.771} & 0.707 & 0.705 & 0.763 & \underline{0.770} \\
\textbf{RTE}   & 0.498 & 0.494 & 0.519 & 0.519 & \underline{\textbf{0.537}} \\
\hline
\textbf{Avg}   & \textbf{0.712} & 0.675 & 0.685 & 0.705 & \underline{0.709} \\
\hline
\end{tabular}
\label{tab:typbert_adding_steps_transposed}
\end{table}

\begin{figure}
    \centering
    \includegraphics[width=0.5\linewidth]{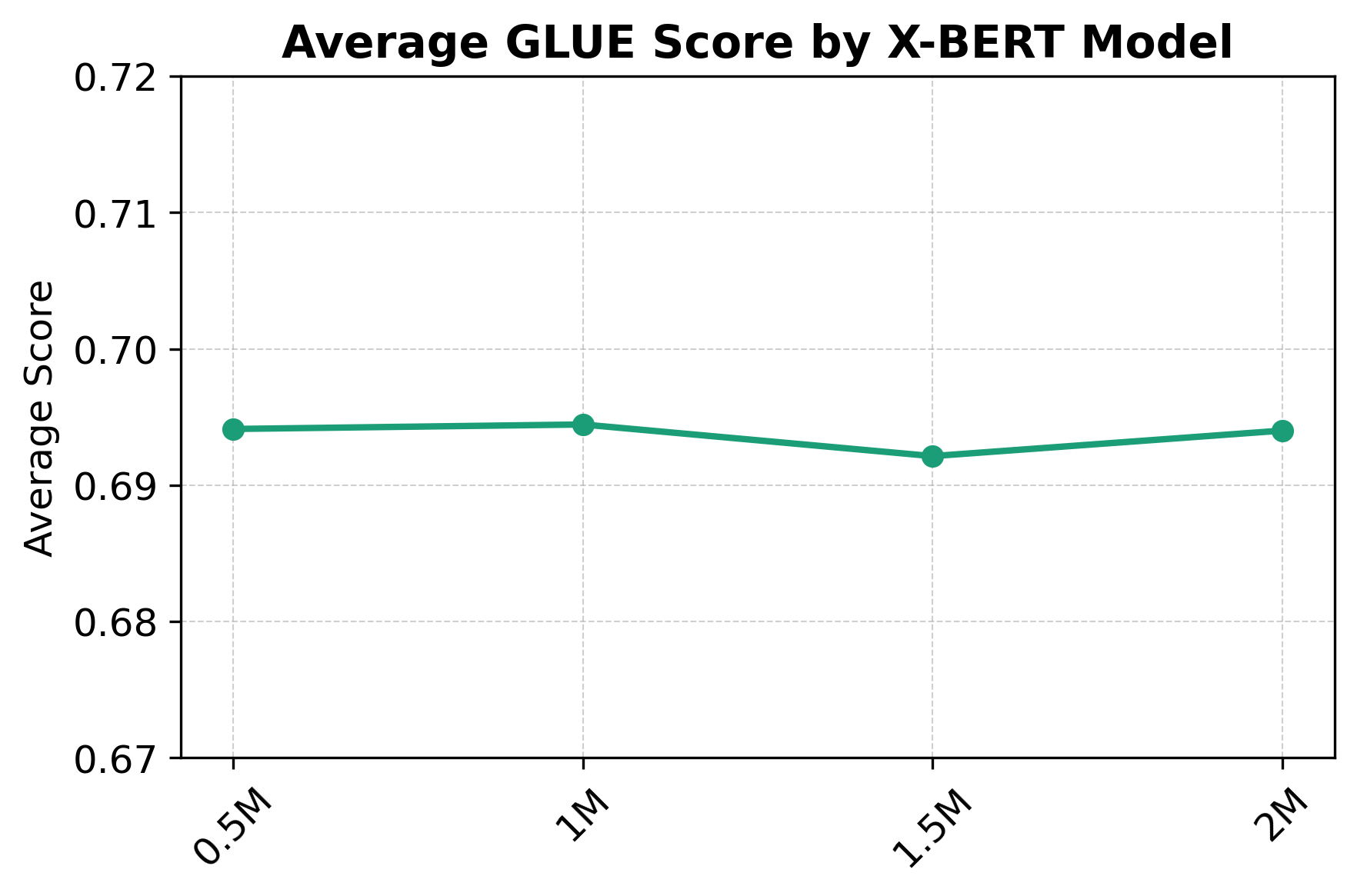}
    \caption{Growth of the average GLUE score of X-BERT variants as training steps increase.}
    \label{fig:XBERT_grown}
\end{figure}

Conversely, X-BERT's performance remained mostly stalled across increased training steps (see Table~\ref{tab:exttypbert_adding_steps_transposed} and Figure~\ref{fig:XBERT_grown}), exhibiting only minor fluctuations and no clear improvement, ultimately being outperformed by T-BERT in later stages. 

\begin{table}[tb]
\centering
\caption{GLUE scores for X-BERT with increasing training steps. Bold indicates the best result overall, while the best results obtained by X-BERT are underlined.}
\begin{tabular}{lccccc}
\hline
Model & C-BERT & X-BERT & & & \\
\hline
\textbf{Task} & \textbf{0.5M} & \textbf{0.5M} & \textbf{1M} & \textbf{1.5M} & \textbf{2M} \\
\hline
\textbf{QQP}   & \textbf{0.852} & 0.844 & 0.842 & 0.846 & \underline{0.848} \\
\textbf{MRPC}  & 0.831 & 0.840 & \underline{\textbf{0.868}} & 0.840 & 0.855 \\
\textbf{CoLA}  & \textbf{0.195} & 0.072 & 0.042 & 0.072 & \underline{0.083} \\
\textbf{QNLI}  & \textbf{0.855} & 0.835 & 0.831 & \underline{0.837} & 0.827 \\
\textbf{STS-B} & \textbf{0.851} & 0.827 & \underline{0.832} & 0.827 & 0.823 \\
\textbf{SST-2} & 0.847 & \textbf{0.858} & 0.839 & 0.842 & 0.844 \\
\textbf{MNLI}  & \textbf{0.771} & 0.754 & \underline{0.754} & 0.750 & 0.742 \\
\textbf{RTE}   & 0.498 & 0.519 & \textbf{0.545} & 0.519 & 0.527 \\
\hline
\textbf{Avg}   & \textbf{0.712} & 0.694 & 0.694 & 0.692 & \underline{0.694} \\
\hline
\end{tabular}
\label{tab:exttypbert_adding_steps_transposed}
\end{table}

This finding is particularly interesting, as the results suggest that there is indeed a loss of information when the entire words are sorted. It is tempting to draw a parallel with the original psycholinguistic observation that humans can still read scrambled text as long as the first and last letters remain in place. While any such analogy between machine and human processing must be made cautiously, our results seem to indicate that preserving the boundary letters may also be a necessary ``anchor'' condition for preserving the distributional regularities that support the distributional hypothesis of meaning \cite{harris1954distributional}.

\section{Conclusion}
\label{sec:conclusions}

\noindent In this study, we investigated typoglycemia in the context of NLP with the aim of better understanding why models trained on typoglycemic text can still achieve competitive performance, despite the linguistic ambiguity introduced by the collapsing effect in English.
Using the British National Corpus, we quantified the frequency and nature of collapsing words and found that such cases are relatively rare and often appear in contexts that clarify their meaning.

We also designed a controlled disambiguation experiment showing that contextual language models (here exemplified by BERT) can use surrounding context to separate collapsing forms in practice.
Building on these observations, we trained three BERT variants—one on clean text, one on text with classic internal character reordering, and one on text with full-word reordering—to analyze the effects of different degrees of typoglycemia on contextual language modeling.
Our results show that models trained on typoglycemic text reach performance levels close to those of models trained on clean text. The same results also show that classic typoglycemia acts as an obfuscation that preserves most of the information (i.e., better performance can be attained by enlarging the training phase), while the extreme variant instead causes a loss of information. 


\section{Limitations and Future Work}
\label{sec:limitations}

\noindent

 This paper has several limitations. First, our analyses were conducted exclusively on English and relying on the British National Corpus, which limits the generability of our conclusions; future work will extend this analysis to other languages
 and different data sources.
 
 Second, we did not test the transferability of typoglycemia to autoregressive architectures or, more generally, to recent LLMs. While our preliminary study focuses on a simple probe study,  extending the experiments to these architectures could enable more well-founded conclusions; something we plan to investigate in future work.
 
 
 Third, our experiments did not systematically vary tokenizers or vocabulary sizes. Since modern transformers rely heavily on tokenization, scrambled inputs may effectively map to arbitrary tokens, particularly when words are represented as single tokens. This could obscure the intended typoglycemia effect, as models may not have access to internal letter-level information. Future work should explore the interaction between tokenization granularity and typoglycemia robustness.

\section*{Acknowledgments}

This work has been supported by the project ``Word Embeddings: From Cognitive Linguistics to Language Engineering, and Back'' (WEMB), funded
by the Italian Ministry of University and Research (MUR) under the PRIN 2022 funding scheme (CUP B53D23013050006).

\clearpage

\appendix

\section{Detailed Accuracy and Error Examples for the Disambiguation Task}
\label{sec:disamb_examples}

\noindent In this section, we present supplementary data from the disambiguation experiment. We report fine-grained accuracy for four target labels under two typoglycemic conditions (classic and extreme), as summarized in Tables~\ref{tab:typoglycemia_detailed_acc} and~\ref{tab:extreme_typoglycemia_detailed_acc}. Each such table shows two cases in which the model achieves almost perfect disambiguation prediction, and two others in which it performs poorly.
As Table~\ref{tab:extreme_typoglycemia_detailed_acc} shows, the extreme typoglycemia condition occasionally produces multiple plausible candidates for disambiguation, as exemplified by the label \emph{there}.

To provide further insight into model behavior, Tables~\ref{tab:sentences_classic_typo} and~\ref{tab:sentences_extreme_typo} present representative examples—covering each label from the accuracy tables—where the model made incorrect predictions. These examples offer a more nuanced view of the experimental outcomes and highlight the challenges posed by typoglycemic perturbations.
Although this remains as a qualitative analysis, we already observe some cases where disambiguation appears challenging even for human annotators.

\section{Tokenization Patterns Across C-BERT, T-BERT, and X-BERT}
\label{sec:more_on_tokenizer}

\noindent In this appendix, we compare the tokenization behavior of C-BERT, T-BERT, and X-BERT.
Table~\ref{tab:top_occurrences_tokenizer} reports the top 20 most frequent tokens for each model variant. 

Overall, token distributions are highly similar across models, with small but notable differences, particularly in the lower-ranked positions.

To further illustrate tokenizer behavior, we selected several frequent\footnote{\url{https://en.wikipedia.org/wiki/Most_common_words_in_English}} English words from Wikipedia (\emph{because, before, people, little, should}) and composed the following test sentence:
\emph{Before people complain, they should remember that little things matter because they build trust.}
We then tokenized this sentence in three versions: the original one, a classic typoglycemia variant, and an extreme typoglycemia variant.
Table~\ref{tab:tokenized_sentence_examples} shows the resulting token sequences, revealing that WordPiece segmentation differs substantially across the three conditions.

\section{Implementation Details and Model Hyperparameters}
\label{sec:model_hyp}

\noindent We conducted all experiments using the \texttt{Transformers} library (v4.50) \cite{wolf2020transformers} for model training and evaluation. Datasets were handled with the \texttt{Datasets} library (v3.3.2). All visualizations were produced using \texttt{matplotlib} (v3.10.1).

For the pretraining of our BERT models, we used a learning rate of $1\times 10^{-4}$ and a weight decay of 0.1.
We did not perform hyperparameter tuning for fine-tuning on GLUE. We adopted hyperparameters following prior work \cite{coco} and other experimental settings, such as\footnote{\url{https://www.tensorflow.org/text/tutorials/bert_glue?utm_source=chatgpt.com}}. We used a weight decay of 0.1, warm-up steps of 1000, and task-specific learning rates as follows: $3\times 10^{-5}$ for tasks MNLI, QNLI, SST-2, STS-B; and $2\times 10^{-5}$ for tasks CoLA, MRPC, QQP, RTE.
Other parameters (e.g., batch size, number of epochs, optimizer settings) are set to their default values in the Hugging Face Transformers library~\cite{wolf2020transformers}.

\section{Additional Tables}
\label{appendix:tables}

\begin{table}[H]
\small
\centering
\caption{Detailed accuracy for some specific labels from the disambiguation experiments for Classic Typoglycemia.}
\begin{tabular}{lll}
\hline
\textbf{label} & \textbf{\#ing-ws} & \textbf{Acc.} \\ \hline
form      & {[}from; form{]}        & 0.997 \\ 
cloud     & {[}could; cloud{]}      & 0.999 \\ 
enquire   & {[}enquire; enrique{]}  & 0.475 \\ 
spectre     & {[}spectre; sceptre{]}      & 0.391 \\ 
\hline
\end{tabular}
\label{tab:typoglycemia_detailed_acc}
\end{table}
\begin{table}[H]
\small
\centering
\caption{Detailed accuracy for some specific labels from the disambiguation experiments for Extreme Typoglycemia.}
\begin{tabular}{lll}
\hline
\textbf{label} & \textbf{\#ing-ws} & \textbf{Acc.} \\ \hline
with       & {[}with; whit{]}                          & 0.999 \\ 
there      & {[}there; three; ether; ehret; reeth{]}   & 0.994 \\ 
presume   & {[}presume; supreme{]}                    & 0.646 \\ 
premiss   & {[}premiss, impress{]}                    & 0.313 \\ \hline
\end{tabular}
\label{tab:extreme_typoglycemia_detailed_acc}
\end{table}

\begin{table}[H]
\small
\centering
\caption{Examples where BERT did not disambiguate correctly in Classic Typoglycemia.}
\begin{minipage}{\textwidth}
\centering
\begin{tabular}{lp{8cm}l}
\hline
\textbf{Golden label} & \textbf{Sentence} & \textbf{Pred.} \\ \hline
form    & \emph{stencil (\textbf{form} your own use card, if desired)} & from \\  
\hline
cloud   & \emph{the ash \textbf{cloud} reach over 15,000 metre in height and is carried by the prevailing wind to the north-east, into Argentina} & could \\
\hline
enquire & \emph{I \textbf{enquire}, as I sit down on the floor beside him.} & Enrique \\ 
\hline
spectre  & \emph{for fairly obvious reason, mostly no doubt connect with the \textbf{spectre} of zhdanov and socialist realism (but also I suspect because of a more subterranean philosophical linkage between callinicos' endorsement of althusser's ‘complex totality’ against Lukács' ‘expressive totality’ and the political voluntarism which it inform in the early twenties), callinicos have to answer: no} & sceptre \\ \hline
\end{tabular}

\label{tab:sentences_classic_typo}
\end{minipage}

\vspace{0.5cm}

\begin{minipage}{\textwidth}
\centering
\caption{Examples where BERT did not disambiguate correctly in Extreme Typoglycemia.}
\begin{tabular}{lp{8cm}l}
\hline
\textbf{Golden label} & \textbf{Sentence} & \textbf{Pred.} \\ \hline
with      & \emph{so it is not as hard as you think w i t h \textbf{with}.} & whit \\ 
\hline
there     & \emph{could you give me your number \textbf{there}?} & three \\ 
\hline
presume   & \emph{\textbf{Presume} guilt is not a selective means of convicting the guilty with more certainty} & supreme \\ 
\hline
premiss   & \emph{For some great conductor, it is — Karajan, like everyone else, has a good line in Knappertsbusch story — but Karajan's repertory and reputation have not been built on that \textbf{premiss}.} & impress \\ \hline
\end{tabular}
\label{tab:sentences_extreme_typo}
\end{minipage}
\end{table}

\begin{table}
\centering
\caption{Top token occurrences for Tokenizer of C-BERT, T-BERT and X-BERT.}
\begin{tabular}{clllrrr}
\hline
\textbf{Pos} & \textbf{C-BERT} & \textbf{T-BERT} & \textbf{X-BERT} & \textbf{Occ. C-BERT} & \textbf{Occ. T-BERT} & \textbf{Occ. X-BERT} \\
\hline
0  & \texttt{[PAD]} & \texttt{[PAD]} & \texttt{[PAD]} & 1224710044 & 1175547259 & 1221126401 \\
1  & \texttt{the} & \texttt{the} & \texttt{the} & 82395952 & 78053447 & 80754785 \\
2  & \texttt{,} & \texttt{,} & \texttt{,} & 74154265 & 70819651 & 73826236 \\
3  & \texttt{.} & \texttt{.} & \texttt{.} & 63910166 & 61122450 & 63650845 \\
4  & \texttt{of} & \texttt{of} & \texttt{of} & 43667581 & 41761752 & 43314058 \\
5  & \texttt{in} & \texttt{in} & \texttt{in} & 41216547 & 39262408 & 40773226 \\
6  & \texttt{and} & \texttt{and} & \texttt{and} & 34490376 & 32489143 & 33723512 \\
7  & \texttt{\#\#s} & \texttt{a} & \texttt{a} & 29422026 & 27865292 & 26436320 \\
8  & \texttt{a} & \texttt{to} & \texttt{to} & 27486517 & 19289061 & 19985593 \\
9  & \texttt{to} & \texttt{\#\#s} & \texttt{-} & 20645743 & 17573839 & 17564619 \\
10 & \texttt{-} & \texttt{-} & \texttt{(} & 17653042 & 16854384 & 17164131 \\
11 & \texttt{(} & \texttt{(} & \texttt{)} & 17247172 & 16585323 & 17090776 \\
12 & \texttt{)} & \texttt{)} & \texttt{was} & 17136023 & 16478979 & 16062162 \\
13 & \texttt{was} & \texttt{was} & \texttt{\#\#s} & 16286474 & 15613034 & 15520095 \\
14 & \texttt{is} & \texttt{is} & \texttt{\#\#u} & 14088556 & 13346144 & 14662273 \\
15 & \texttt{"} & \texttt{\#\#t} & \texttt{\#\#t} & 11114076 & 10853839 & 14588935 \\
16 & \texttt{\#\#e} & \texttt{\#\#e} & \texttt{is} & 10775381 & 10587162 & 13670252 \\
17 & \texttt{as} & \texttt{"} & \texttt{\#\#y} & 10627122 & 10467354 & 13153451 \\
18 & \texttt{for} & \texttt{as} & \texttt{"} & 9895493  & 9976799  & 10997527 \\
19 & \texttt{on} & \texttt{\#\#n} & \texttt{as} & 9847865  & 9866984  & 10641475 \\
20 & \texttt{he} & \texttt{\#\#a} & \texttt{he} & 9618214  & 9367440  & 9903201 \\
\hline
\end{tabular}
\label{tab:top_occurrences_tokenizer}
\end{table}

\begin{table}
\small
\centering
\caption{Exact token sequences from Tokenizers C-BERT, T-BERT, and X-BERT in Python list format. Unaltered terms for T-BERT and X-BERT correspond to words that are taken as complete tokens by WordPiece.}
\resizebox{0.75\columnwidth}{!}{
\begin{tabular}{p{6cm}cp{4cm}}
\hline
\textbf{Sentence} & \textbf{Tokenizer} & \textbf{Tokenized Sentence} \\
\hline
Before people complain they should remember that little things matter because they build trust
& C-BERT
& \texttt{['before', 'people', 'compl', '\#\#ain', 'they', 'should', 'rem', '\#\#ember', 'that', 'little', 'things', 'matter', 'because', 'they', 'build', 'trust']} \\
\hline
Before pelope cailmopn tehy shloud rbeeemmr taht liltte tghins maettr bacesue tehy bilud trsut 
& T-BERT
& \texttt{['before', 'pelope', 'cailm', '\#\#opn', 'tehy', 'shloud', 'rbee', '\#\#em', '\#\#mr', 'taht', 'liltte', 'tghins', 'mae', '\#\#ttr', 'bacesue', 'tehy', 'bilud', 't', '\#\#rsut']} \\
\hline
Beefor eeelop acilmnop ehty dhloos beeemmrr ahtt eilltt ghinst aemrtt abceesu ehty bdilu rsttu
& X-BERT
& \texttt{['beefor', 'eee', '\#\#lop', 'acilm', '\#\#nop', 'ehty', 'dhlo', '\#\#os', 'beee', '\#\#mmr', '\#\#r', 'ahtt', 'eilltt', 'ghinst', 'aem', '\#\#rtt', 'abceesu', 'ehty', 'bdilu', 'r', '\#\#sttu']} \\
\hline
\end{tabular}}
\label{tab:tokenized_sentence_examples}
\end{table}

\clearpage
\bibliographystyle{apalike}  
\bibliography{references}  

\begin{thebibliography}{}

\bibitem[Andrews, 1996]{andrews1996lexical}
Andrews, S. (1996).
\newblock Lexical retrieval and selection processes: Effects of transposed-letter confusability.
\newblock {\em Journal of Memory and Language}, 35(6):775--800.

\bibitem[Belinkov and Bisk, 2018]{belinkovB18}
Belinkov, Y. and Bisk, Y. (2018).
\newblock Synthetic and natural noise both break neural machine translation.
\newblock In {\em Proceedings of the 6th International Conference on Learning Representations (ICLR 2018)}, Vancouver, CA.

\bibitem[Brezina and Meyerhoff, 2014]{significant_or_random}
Brezina, V. and Meyerhoff, M. (2014).
\newblock Significant or random?: A critical review of sociolinguistic generalisations based on large corpora.
\newblock {\em International Journal of Corpus Linguistics}, 19.

\bibitem[Cao et~al., 2023]{CaoKMI23}
Cao, Q., Kojima, T., Matsuo, Y., and Iwasawa, Y. (2023).
\newblock Unnatural error correction: {GPT-4} can almost perfectly handle unnatural scrambled text.
\newblock In Bouamor, H., Pino, J., and Bali, K., editors, {\em Proceedings of the 2023 Conference on Empirical Methods in Natural Language Processing, {EMNLP} 2023, Singapore, December 6-10, 2023}, pages 8898--8913. Association for Computational Linguistics.

\bibitem[Devlin et~al., 2019]{bert}
Devlin, J., Chang, M., Lee, K., and Toutanova, K. (2019).
\newblock {BERT: P}re-training of deep bidirectional transformers for language understanding.
\newblock In {\em Proceedings of the 2019 Conference of the North American Chapter of the Association for Computational Linguistics (NAACL 2019)}, pages 4171--4186, Minneapolis, {US}.

\bibitem[Grainger and Whitney, 2004]{grainger2004does}
Grainger, J. and Whitney, C. (2004).
\newblock Does the huamn mnid raed wrods as a wlohe?
\newblock {\em Trends in cognitive sciences}, 8(2):58--59.

\bibitem[Harris, 1954]{harris1954distributional}
Harris, Z.~S. (1954).
\newblock Distributional structure.
\newblock {\em Word}, 10(2-3):146--162.

\bibitem[Healy, 1976]{healy1976detection}
Healy, A.~F. (1976).
\newblock Detection errors on the word the: Evidence for reading units larger than letters.
\newblock {\em Journal of Experimental Psychology: Human Perception and Performance}, 2(2):235.

\bibitem[Heigold et~al., 2018]{heigold-etal-2018-robust}
Heigold, G., Varanasi, S., Neumann, G., and van Genabith, J. (2018).
\newblock How robust are character-based word embeddings in tagging and {MT} against wrod scramlbing or randdm nouse?
\newblock In {\em Proceedings of the 13th Conference of the Association for Machine Translation in the {A}mericas (Volume 1: Research Track)}, pages 68--80, Boston, US.

\bibitem[Kim, 2014]{Kim14charcnn}
Kim, Y. (2014).
\newblock Convolutional neural networks for sentence classification.
\newblock In {\em Proceedings of the 2014 Conference on Empirical Methods in Natural Language Processing, (EMNLP 2014)}, pages 1746--1751, Doha, QA.

\bibitem[Kumar et~al., 2020]{kumarMG20}
Kumar, A., Makhija, P., and Gupta, A. (2020).
\newblock Noisy text data: {A}chilles' heel of {BERT}.
\newblock In {\em Proceedings of the 6th Workshop on Noisy User-Generated Text (NUT 2020)}, pages 16--21, Online Event.

\bibitem[Malykh et~al., 2018]{MalykhLK18}
Malykh, V., Logacheva, V., and Khakhulin, T. (2018).
\newblock Robust word vectors: {C}ontext-informed embeddings for noisy texts.
\newblock In {\em Proceedings of the 4th Workshop on Noisy User-Generated Text (NUT 2018)}, pages 54--63, Brussels, BE.

\bibitem[Marian et~al., 2012]{marian2012clearpond}
Marian, V., Bartolotti, J., Chabal, S., and Shook, A. (2012).
\newblock Clearpond: Cross-linguistic easy-access resource for phonological and orthographic neighborhood densities.
\newblock {\em PLOS ONE}, 7(8):e43230.

\bibitem[McCusker et~al., 1981]{mccusker1981word}
McCusker, L.~X., Gough, P.~B., and Bias, R.~G. (1981).
\newblock Word recognition inside out and outside in.
\newblock {\em Journal of Experimental Psychology: Human Perception and Performance}, 7(3):538.

\bibitem[Meng et~al., 2021]{coco}
Meng, Y., Xiong, C., Baja, P., Tiwary, S., Bennett, P., Han, J., and Song, X. (2021).
\newblock Coco-lm: correcting and contrasting text sequences for language model pretraining.
\newblock In {\em Proceedings of the 35th International Conference on Neural Information Processing Systems}, NIPS '21, Red Hook, NY, US. Curran Associates Inc.

\bibitem[Moradi and Samwald, 2021]{moradi}
Moradi, M. and Samwald, M. (2021).
\newblock Evaluating the robustness of neural language models to input perturbations.
\newblock In {\em Proceedings of the 2021 Conference on Empirical Methods in Natural Language Processing (EMNLP 2021)}, pages 1558--1570, Punta Cana, DO.

\bibitem[Nguyen and Grieve, 2020]{nguyen2020word}
Nguyen, D. and Grieve, J. (2020).
\newblock Do word embeddings capture spelling variation?
\newblock In {\em Proceedings of the 28th International Conference on Computational Linguistics}, pages 870--881, Barcelona, ES.

\bibitem[OpenAI et~al., 2024]{gpt4}
OpenAI, Achiam, J., and {283 other authors} (2024).
\newblock {GPT}-4 technical report.
\newblock arXiv:2303.08774 [cs.CL].

\bibitem[Pan et~al., 2024]{panLX24}
Pan, L., Leng, Y., and Xiong, D. (2024).
\newblock Can large language models learn translation robustness from noisy-source in-context demonstrations?
\newblock In {\em Proceedings of the 2024 Joint International Conference on Computational Linguistics, Language Resources and Evaluation, {LREC/COLING} 2024, 20-25 May, 2024, Torino, Italy}, pages 2798--2808. {ELRA} and {ICCL}.

\bibitem[Pruthi et~al., 2019]{PruthiDL19}
Pruthi, D., Dhingra, B., and Lipton, Z.~C. (2019).
\newblock Combating adversarial misspellings with robust word recognition.
\newblock In {\em Proceedings of the 57th Conference of the Association for Computational Linguistics (ACL 2019)}, pages 54--63, Firenze, {IT}.

\bibitem[Ravichander et~al., 2021]{RavichanderDRMH21}
Ravichander, A., Dalmia, S., Ryskina, M., Metze, F., Hovy, E.~H., and Black, A.~W. (2021).
\newblock {NoiseQA: C}hallenge set evaluation for user-centric question answering.
\newblock In {\em Proceedings of the 16th Conference of the European Chapter of the Association for Computational Linguistics (EACL 2021)}, pages 2976--2992, Online Event.

\bibitem[Rayner et~al., 2006]{jumbledwords}
Rayner, K., White, S., Johnson, R., and Liversedge, S. (2006).
\newblock Raeding wrods with jubmled lettres: There is a cost.
\newblock {\em Psychological Science}, 17:192--3.

\bibitem[Sakaguchi et~al., 2017]{SakaguchiDPD17}
Sakaguchi, K., Duh, K., Post, M., and Durme, B.~V. (2017).
\newblock Robsut wrod reocginiton via semi-character recurrent neural network.
\newblock In {\em Proceedings of the 31st Conference on Artificial Intelligence (AAAI 2017)}, pages 3281--3287, San Francisco, US.

\bibitem[Satheesh et~al., 2025]{satheesh}
Satheesh, S., Beckh, K., Klug, K., Allende{-}Cid, H., Houben, S., and Hassan, T. (2025).
\newblock Robustness evaluation of the {G}erman extractive question answering task.
\newblock In {\em Proceedings of the 31st International Conference on Computational Linguistics, {COLING} 2025, Abu Dhabi, UAE, January 19-24, 2025}, pages 1785--1801. Association for Computational Linguistics.

\bibitem[Sperduti et~al., 2021]{sperduti2021garbled}
Sperduti, G., Moreo, A., and Sebastiani, F. (2021).
\newblock Garbled-word embeddings for jumbled text.
\newblock In {\em IIR}.

\bibitem[{The British National Corpus Consortium}, 2007]{bnc2007}
{The British National Corpus Consortium} (2007).
\newblock The british national corpus, version 3 (bnc xml edition).
\newblock Distributed by Oxford University Computing Services on behalf of the BNC Consortium.
\newblock Available at \url{http://www.natcorp.ox.ac.uk/}.

\bibitem[Wang et~al., 2018]{glue}
Wang, A., Singh, A., Michael, J., Hill, F., Levy, O., and Bowman, S.~R. (2018).
\newblock {GLUE:} {A} multi-task benchmark and analysis platform for natural language understanding.
\newblock In Linzen, T., Chrupala, G., and Alishahi, A., editors, {\em Proceedings of the Workshop: Analyzing and Interpreting Neural Networks for NLP, BlackboxNLP@EMNLP 2018, Brussels, Belgium, November 1, 2018}, pages 353--355. Association for Computational Linguistics.

\bibitem[Wang et~al., 2025]{wang2025}
Wang, C., Gu, T., Wei, Z., Gao, L., Song, Z., and Chen, X. (2025).
\newblock Word form matters: Llms' semantic reconstruction under typoglycemia.
\newblock In Che, W., Nabende, J., Shutova, E., and Pilehvar, M.~T., editors, {\em Findings of the Association for Computational Linguistics, {ACL} 2025, Vienna, Austria, July 27 - August 1, 2025}, pages 16870--16885. Association for Computational Linguistics.

\bibitem[Warstadt et~al., 2018]{cola}
Warstadt, A., Singh, A., and Bowman, S.~R. (2018).
\newblock Neural network acceptability judgments.
\newblock {\em arXiv preprint arXiv:1805.12471}.

\bibitem[Wolf et~al., 2020]{wolf2020transformers}
Wolf, T., Debut, L., Sanh, V., Chaumond, J., Delangue, C., Moi, A., Cistac, P., Rault, T., Louf, R., Funtowicz, M., Davison, J., Shleifer, S., von Platen, P., Ma, C., Jernite, Y., Plu, J., Xu, C., Le~Scao, T., Gugger, S., Drame, M., Lhoest, Q., and Rush, A. (2020).
\newblock Transformers: State-of-the-art natural language processing.
\newblock In Liu, Q. and Schlangen, D., editors, {\em Proceedings of the 2020 Conference on Empirical Methods in Natural Language Processing: System Demonstrations}, pages 38--45, Online. Association for Computational Linguistics.

\bibitem[Wulff, 2003]{adjective_english}
Wulff, S. (2003).
\newblock A multifactorial corpus analysis of adjective order in english.
\newblock {\em International Journal of Corpus Linguistics}, 8:245--282.

\bibitem[Yang and Gao, 2019]{yang2019can}
Yang, R. and Gao, Z. (2019).
\newblock Can machines read jmulbed senetcnes?

\bibitem[Zhang et~al., 2025]{ebench}
Zhang, Z., Hao, B., Li, J., Zhang, Z., and Zhao, D. (2025).
\newblock {E}-bench: Towards evaluating the ease-of-use of large language models.
\newblock In Rambow, O., Wanner, L., Apidianaki, M., Al-Khalifa, H., Eugenio, B.~D., and Schockaert, S., editors, {\em Proceedings of the 31st International Conference on Computational Linguistics}, pages 2329--2339, Abu Dhabi, UAE. Association for Computational Linguistics.

\end{thebibliography}

\end{document}